\documentclass[journal,article,submit,moreauthors,pdftex]{Definitions/mdpi} 

\firstpage{1} 
\pubvolume{xx}
\issuenum{1}
\articlenumber{5}
\pubyear{2020}
\copyrightyear{2020}
\history{Received: date; Accepted: date; Published: date}

\usepackage{url}            
\usepackage{booktabs}       

\usepackage{multirow}
\usepackage{textcomp}

\usepackage{graphicx}
\usepackage{caption}
\usepackage{subcaption}

\usepackage{float}

\usepackage{amsmath}
\usepackage{longtable,booktabs,multicol,multirow,lineno}

\usepackage{hyperref}       

\newcommand{\eg}{\emph{e.g.}}
\newcommand{\ie}{\emph{i.e.}}

\Title{Robust Representation and Efficient Feature Selection Allows
  for Effective Clustering of SARS-CoV-2 Variants}

\Author{Zahra Tayebi $^{1}$, Sarwan Ali $^{2}$ and Murray Patterson $^{3,}$*}

\AuthorNames{Zahra Tayebi, Sarwan Ali and Murray Patterson}

\address{%
$^{1}$ \quad Georgia State University, Atlanta, GA, USA; ztayebi1@student.gsu.edu \\
$^{2}$ \quad Georgia State University, Atlanta, GA, USA; sali85@student.gsu.edu \\
$^{3}$ \quad Georgia State University, Atlanta, GA, USA; mpatterson30@gsu.edu
}

\corres{Correspondence: mpatterson30@gsu.edu}

\abstract{%
  The widespread availability of large amounts of genomic data on the
  SARS-CoV-2 virus, as a result of the COVID-19 pandemic, has created
  an opportunity for researchers to analyze the disease at a level of
  detail unlike any virus before it.  One one had, this will help
  biologists, policy makers and other authorities to make timely and
  appropriate decisions to control the spread of the coronavirus.  On
  the other hand, such studies will help to more effectively deal with
  any possible future pandemic.  Since the SARS-CoV-2 virus contains
  different variants, each of them having different mutations,
  performing any analysis on such data becomes a difficult task.  It
  is well known that much of the variation in the SARS-CoV-2 genome
  happens disproportionately in the spike region of the genome
  sequence --- the relatively short region which codes for the spike
  protein(s).  Hence, in this paper, we propose an approach to cluster
  spike protein sequences in order to study the behavior of different
  known variants that are increasing at very high rate throughout the
  world.  We use a k-mers based approach to first generate a
  fixed-length feature vector representation for the spike sequences.
  We then show that with the appropriate feature selection, we can
  efficiently and effectively cluster the spike sequences based on the
  different variants.  Using a publicly available set of SARS-CoV-2
  spike sequences, we perform clustering of these sequences using both
  hard and soft clustering methods and show that with our feature
  selection methods, we can achieve higher F1 scores for the clusters.
}


\begin{document}









\section{Introduction}

The virus that causes the COVID-19 disease is called the severe acute
respiratory syndrome coronavirus 2 (SARS-CoV-2) --- a virus whose
genomic sequence is being replicated and dispersed across the globe at
an extraordinary rate.  The genomic sequences of a virus can be
helpful to investigate outbreak dynamics such as spatiotemporal
spread, the size variations of the epidemic over time, and
transmission routes.  Furthermore, genomic sequences can help design
investigative analyses, drugs and vaccines, and monitor whether
theoretical changes in their effectiveness over time might refer to
changes in the viral genome.  Analysis of SARS-CoV-2 genomes can
therefore complement, enhance and support strategies to reduce the
burden of COVID-19~\cite{world2021genomic}.

\begin{figure}
  \centering
  \includegraphics[scale = 0.52] {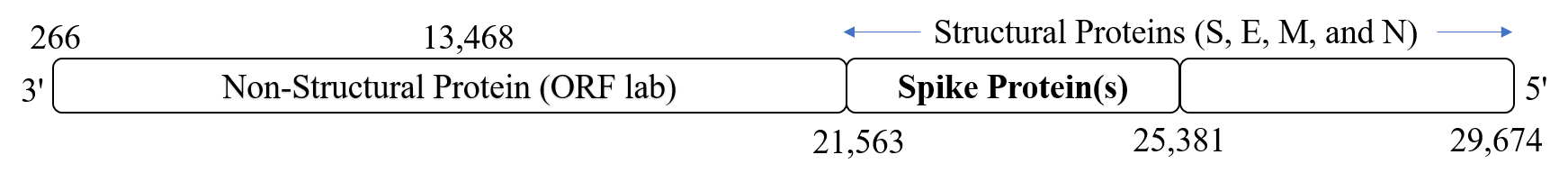}
  \caption{The SARS-CoV-2 genome is roughly 29--30kb in length,
    encoding structural and non-structural proteins.  Open reading
    frame (ORF) 1ab encodes the non-structural proteins, and the four
    structural proteins: S (spike), E (envelope), M (membrane), and N
    (nucleocapsid) are encoded by their respective genes.  The spike
    protein has $1274$ amino acids.}
  \label{fig_spikeprot}
\end{figure}

SARS-CoV-2 is a single-stranded RNA-enveloped
virus~\cite{lu2020genomic}.  Its entire genome is characterized by
applying an RNA-based metagenomic next-generation sequencing method.
The length of the genome is 29,881 bp (GenBank no. MN908947), encoding
9860 amino acids \cite{chen2020rna}.  Structural and nonstructural
proteins are expressing the gene fragments.  Structural proteins are
encoded by the S, E, M and N genes, while the ORF region encodes
nonstructural proteins \cite{chan2020genomic} (see
Figure~\ref{fig_spikeprot}).

A key factor involved in infection is the S protein on the surface of
the virus~\cite{weissenhorn1999structural}.  The S protein of
SARS-CoV-2 is similar to other coronaviruses and arbitrates receptor
recognition, fusion, and cell attachment through viral
infection~\cite{walls2020structure,lan2020structure,gui2017cryo}.  The
S protein has an essential role in viral infection that makes it a
potential target for vaccine development, antibody-blocking therapy,
and small molecule inhibitors~\cite{huang2020structural}.  Also, the
spike region of the SARS-CoV-2 genome is involved in a
disproportionate amount of the genomic variation, for its
length~\cite{kuzmin2020machine} (see, \eg,
Table~\ref{tbl_variant_information}).  Therefore, mutations that
affect the antigenicity of the S protein are of certain
importance~\cite{harvey2021sars}.

Generally, the genetic variations of a virus are grouped into clades,
which can also be called subtypes, genotypes, or groups.  To study the
evolutionary dynamics of viruses, building pylogenetic trees out of
sequences is common~\cite{hadfield2018a}.  On the other hand, the
number of available SARS-CoV-2 sequences is huge and still
increasing~\cite{guyon2003introduction} --- building trees on the
millions of SARS-CoV-2 sequences would be very expensive and seems
impractical.  In these cases, machine learning approaches that have
flexibility and scalability could be useful~\cite{ngiam2019big}.
Since natural clusters of the sequences are formed by the major
clades, clustering methods would be useful to understand the
complexity behind the spread of the COVID-19 in terms of its
variation.  Also by considering the certain importance of the S
protein, we focus on the amino acid (protein) sequences encoded by the
spike region.  In this way, we would reduce the dimensionality of data
without losing too much information, reducing the time and storage
space required and making visualization of the data
easier~\cite{van2009dimensionality}.

\begin{figure}
  \centering \includegraphics[scale = 0.95] {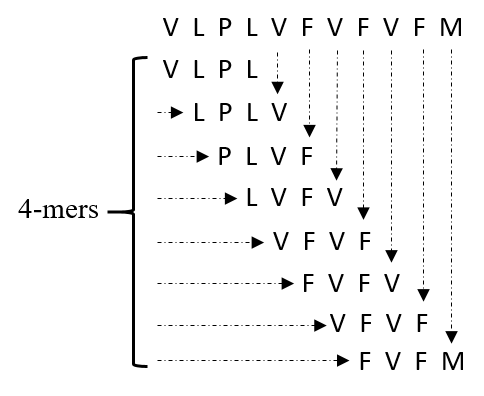}
  \caption{Example of 4-mers of the amino acid sequence
    ``VLPLVFVFVFM''.}
  \label{fig_k_mer_demo}
\end{figure}

To make use of machine learning approaches, we need to prepare the
appropriate input --- numerical (real-valued) vectors --- that is
compatible with these methods.  This would give us the ability to
perform meaningful analytics.  As a result, these amino acid sequences
should be converted into numeric characters in a way that preserves
some sequential order information of the amino acids within each
sequence.  The most prevalent strategy in this area is one-hot
encoding due to its simplicity~\cite{kuzmin2020machine}.  Since we
need to compute pairwise distances (\eg, Euclidean distance), one-hot
encoding order preservation would not be operational~\cite{ali2021k}.
To preserve order information of each sequence while being amenable to
pairwise distance computation, $k$-mers (length $k$ substrings of each
sequence) are calculated and input to the downstream
classification/clustering tasks~\cite{ali2021k,farhan2017efficient}
(see Figure~\ref{fig_k_mer_demo}).

The proposed methods in this study are fast and efficient clustering
methods to cluster the spike amino acid sequences of SARS-CoV-2.  We
demonstrate that our method performs considerably better than the
basic methods, and the variants are successfully clustered into unique
clusters with high $F_1$ score.  The following are the contributions
of this paper:
\begin{enumerate}
\item For efficient sequence clustering, we propose a method based on
  $k$-mers, and show that the downstream clustering methods
  successfully cluster the variants with high $F_1$ score.
\item We performed experiments using different clustering algorithms
  and feature selection approaches and show the trad-off between the
  clustering quality and the runtime for these methods.
\item We use both hard and soft clustering approaches to study the
  behavior of different coronavirus variants in detail.
\end{enumerate}

The rest of the paper is organized as follows:
Section~\ref{related_work} contains related work of our approach.  Our
proposed approach is detailed in Section~\ref{sec_proposed_approach}.
A description of the datasets used are given in
Section~\ref{sec_experimental_setup}.  We provide a detailed
discussion about the results in Section~\ref{sec_results_discussion},
and then we conclude our paper in Section~\ref{sec_conclusion}.



\section{Literature Review}
\label{related_work}

Performing different data analytics tasks on sequences has been done
successfully by different researchers
previously~\cite{Krishnan2021PredictingVaccineHesitancy,ali2021k}. However,
most studies require the sequences to be
aligned~\cite{Dwivedi2012ClassificationOH,kuzmin2020machine,melnyk-2021-alpha}.
The aligned sequences are used to generate fixed length numerical
embeddings, which can then used for tasks such as classification and
clustering~\cite{ali2021k,ali2021effective,ali2021spike2vec}. Since
the dimensionality of data is another problem while dealing with
larger sized sequences, using approximate methods to compute the
similarity between two sequences is a popular
approach~\cite{Kuksa_SequenceKernel,hoffmann2007kernel,farhan2017efficient}.
The fixed-length numerical embedding methods have been successfully
used in literature for other applications such as predicting missing
values in graphs~\cite{ali2021predicting}, text
analytics~\cite{shakeel2019multi,shakeel2020language,shakeel2020multi},
biology~\cite{leslie2002mismatch,farhan2017efficient,Kuksa_SequenceKernel},
graph analytics~\cite{hassan2020estimating,Hassan2021Computing},
classification of electroencephalography and electromyography
sequences~\cite{atzori2014electromyography,ullah2020effect}, detecting
security attacks in networks~\cite{Ali2019Detecting1}, and electricity
consumption in smart grids~\cite{ali2019short,Ali2020ShortTerm}.  The
conditional dependencies between variables is also important to study
so that their importance can be analyzed in
detail~\cite{ali2021cache}.

Due to the availability of large-scale sequence data for the
SARS-CoV-2 virus, an accurate and effective clustering method is
needed to further analyze this disease, so as to better understand the
dynamics and diversity of this virus. To classify different
coronavirus hosts, authors in~\cite{kuzmin2020machine} suggest a
one-hot encoding-based method that uses spike sequences alone. Their
study reveals that they achieved excellent prediction accuracy
considering just the spike portion of the genome sequence instead of
using the entire sequence. Using this idea and a kernel method, Ali et
al., in~\cite{ali2021k} accomplish higher accuracy than
in~\cite{kuzmin2020machine}, in classification of different variants
of SARS-CoV-2 in humans. Successfully analysis of different variants
leads to designing efficient strategy regarding the vaccination
distribution~\cite{Ahmad2016AusDM,AHMAD2020Combinatorial,ahmad2017spectral,Tariq2017Scalable}.

\section{Proposed Approach}\label{sec_proposed_approach}
In this section, we discuss our proposed algorithm in detail. We start with the description of $k$-mers generation from the spike sequences. We then describe how we generated the feature vector representation from the $k$-mers information. After that, we discuss different feature selection methods in detail. Finally, we detail how we applied clustering approaches on the final feature vector representation.

\subsection{k-mers Generation}
Given a spike sequence, the first step is to compute all possible $k$-mers. The total number of $k$-mers that we can generate for a spike sequence are described as follows:
\begin{equation}
    N - k + 1
\end{equation}
where $N$ is the length of the spike sequence ($N = 1274$ for our
dataset). The variable $k$ is a user-defined parameter (we took $k =
3$ using standard validation set
approach~\cite{validationSetApproach}). For an example of how to
generate $k$-mers, see Figure~\ref{fig_k_mer_demo}.

\subsection{Fixed-Length Feature Vector Generation}
Since most of the Machine Learning (ML) models work with a fixed-length feature vector representation, we need to convert the $k$-mers information into the vectors. For this purpose, we generate a feature vector $\Phi_k$ for a given spike sequence $a$ (\ie, $\Phi_k(a)$). Given an alphabet $\Sigma$ (characters representing amino acids in the spike sequence), the length of $\Phi_k(a)$ will be equal to the number of possible $k$-mers of $a$. More formally,
\begin{equation}
    \Phi_k(a) = \vert \Sigma \vert^k
\end{equation}
Since we have $21$ unique characters in $\Sigma$ (namely \textit{ACDEFGHIKLMNPQRSTVWXY}), the length of each frequency vector is $21^3 = 9261$.

\subsection{Low Dimensional Representation}
Since the dimensionality of data is high after getting the fixed length feature vector representation, we apply different supervised and unsupervised methods to obtain a low dimensional representation of data to avoid the problem of the \textit{curse of dimensionality}~\cite{ali2019short,ali2019short_AMI}. Each of the methods for obtaining a low dimensional representation of data is discussed below:

\subsubsection{Random Fourier Features}
The first method that we use is an approximate kernel method called Random Fourier Features (RFF)~\cite{rahimi2007random}. It is an unsupervised approach, which maps the input data to a
randomized low dimensional feature space (euclidean inner product
space) to get an approximate representation of data in lower dimensions $D$ from the original dimensions $d$. More formally:
\begin{equation}
  z: \mathcal{R}^d \rightarrow \mathcal{R}^D
\end{equation}
In this way, we approximate the inner product between a pair of
transformed points. More formally:
\begin{equation}\label{eq_z_value}
  f(x,y) = \langle \phi (x), \phi (y) \rangle \approx z(x)' z(y)
\end{equation}
In Equation~\eqref{eq_z_value}, $z$ is low dimensional (unlike the
lifting $\phi$). 
Now, $z$ acts as the approximate low dimensional embedding for
the original data. We can use $z$ as
an input for different ML tasks like clustering and classification.

\subsubsection{Least Absolute Shrinkage and Selection Operator (Lasso) Regression}
Lasso regression is a supervised method that can be used for efficient feature selection. It is a type of regularized linear regression variants. 
It is a specific case of the penalized least squares regression with an $L_1$ penalty function. By combining the good qualities of ridge regression~\cite{hoerl1975ridge,mcdonald2009ridge} and subset selection, Lasso can improve both model interpretability and prediction accuracy \cite{muthukrishnan2016lasso}. Lasso regression tries to minimize the following objective function:
\begin{equation}
    min(\text{Sum of square residuals} + \alpha \times \vert \text{slope} \vert)
\end{equation}
where $\alpha \times \vert \text{slope} \vert$ is the penalty term. In Lasso regression, we take the absolute value of the slope in the penalty term rather than the square (as in ridge regression~\cite{mcdonald2009ridge}). This helps to reduce the slope of useless variables exactly equal to zero.

\subsubsection{Boruta}
The last feature selection method that we are using is Boruta. It is a supervised method that is made all around the random forest (RF) classification algorithm. It works by creating shadow features so that the features do not compete among themselves but rather they compete with a randomized version of them~\cite{kursa2010feature}. 
It captures the non-linear relationships and interactions using the RF algorithm. It then extract the importance of each feature (corresponding to the class label) and only keep the features that are above a specific threshold of importance. The threshold is defined as the highest feature importance recorded among the shadow features.


\subsection{Clustering Methods}
In this paper, we use five different clustering methods (both hard and soft clustering approaches) namely k-means~\cite{fahim2006efficient}, k-modes~\cite{khan2013cluster}, Fuzzy c-means~\cite{bezdek1984fcm,dias2019fuzzy}, agglomerative hierarchical clustering, and  Hierarchical density-based spatial clustering of applications with noise (HDBSCAN)~\cite{campello2013density,mcinnes2017hdbscan} (note that is is a soft clustering approach). For the k-means and k-modes, default parameters are used. For the fuzzy c-means, the clustering criterion used to aggregate subsets is a generalized least-squares objective function. For agglomerative hierarchical clustering, a bottom-up approach is applied, which is acknowledged as the agglomerative method. Since the bottom-up procedure starts from anywhere in the central point of the hierarchy and the lower part of the hierarchy is developed by a less expensive method such as partitional clustering, it can reduce the computational cost \cite{bouguettaya2015efficient}.

HDBSCAN is not just density-based spatial clustering of applications with noise (DBSCAN) but switching it into a hierarchical clustering algorithm and then obtaining a flat clustering based in the solidity of clusters.
HDBSCAN is robust to parameter choice and can discover clusters of differing densities (unlike DBSCAN) \cite{mcinnes2017hdbscan}.

\subsection{Optimal number of Clusters}
\begin{figure}
    \centering
    \includegraphics[scale = 0.55]{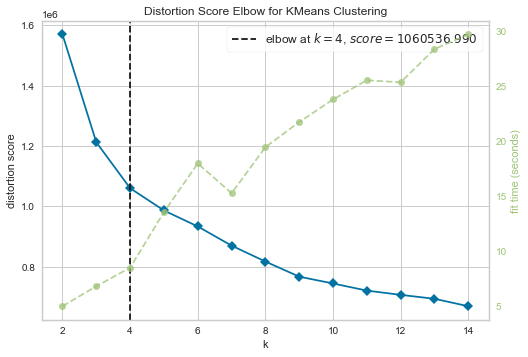}
    \caption{The distortion score (blue line) for different numbers of
      clusters using $k$-means.  The dashed green line shows the
      runtime (in sec.).  The dashed black line shows the optimal
      number of clusters computed using the Elbow
      method~\cite{satopaa2011finding}.}
    \label{fig_ideal_k_value}
\end{figure}

We determined the optimal number of clusters using the elbow method~\cite{satopaa2011finding}. It can fit the model with number of clusters K ranging from $2$ to $14$. As a quality measure, ‘distortion’ is used, which measures the sum of squared distances from each point to its center. Figure~\ref{fig_ideal_k_value} is showing the distortion score for several values of K. We also plot the training runtime (in seconds) to see the trade-off between distortion score  and the runtime. We use the “knee point detection algorithm (KPDA)"~\cite{satopaa2011finding} to determine the optimal number of clusters. Note that based on results shown in Figure~\ref{fig_ideal_k_value}, the perfect number of clusters is $4$. However, we choose K = 5 for all hard clustering approaches because we have five different variants in our data (see Table~\ref{tbl_variant_information}).
The KPDA chose four as the best initial number of clusters due to the Beta variant being not well-represented in the data (see Table~\ref{tbl_variant_information}). However, to give a fair chance to the Beta variant to form its own cluster, we choose $5$ as the number of clusters.

\section{Experimental Setup}\label{sec_experimental_setup}

In this section, first, we provide information associated to the dataset. Then, with the benefit of the $t$-distributed stochastic neighbor embedding ($t$-SNE)~\cite{van2008visualizing}, we try to reduce dimensions with non-linear relationships to find any natural hidden clustering in the data. This data analysis step helps us to obtain basic knowledge about different variants.
As a baseline, we use $k$-mers based frequency vectors without applying any feature selection to perform clustering using k-means, k-modes, fuzzy, hierarchical, and Density-based spatial (HDBSCAN) algorithms.
The weighted $F_1$ score is used to measure the quality of clustering algorithms for different experimental settings.
All experiments are performed on a Core i5 system running the Windows operating system, 32GB memory, and a 2.4 GHz processor. Implementation of the algorithms is done in Python, and the code is available online\footnote{\url{https://github.com/sarwanpasha/COVID_19_Community_Detection_For_Variants/tree/main/Results}}. Our
pre-processed data is also available
online\footnote{\url{https://drive.google.com/drive/folders/1-YmIM8ipFpj-glr9hSF3t6VuofrpgWUa?usp=sharing}},
which can be used after agreeing to terms and conditions of
GISAID\footnote{\url{https://www.gisaid.org/}}. The code of HDBSCAN is taken from~\cite{mcinnes2017hdbscan}. The code for fuzzy c-means is also available online\footnote{\url{https://github.com/omadson/fuzzy-c-means}}.

\subsection{Dataset Statistics}
Our dataset is the (aligned) amino acid sequences (spike region only) of the SARS-CoV-2 proteome. The dataset is publicly available on the GISAID website\footnote{\url{https://www.gisaid.org/}}, which is the largest known database of SARS-CoV-2 sequences. Table~\ref{tbl_variant_information} is showing more information related to the dataset.
There are five most common variants namely Alpha, Beta, Delta, Gamma, and Epsilon. The forth column of Table~\ref{tbl_variant_information} shows number of mutations occurred in spike protein over the number of total mutations (in whole genome) for each variant, e.g,. for Alpha variant there are 17 mutations in the whole genome and 8  of the mutations are in spike region out of those 17. In our dataset, we have 62,657 amino acid sequences (after removing missing values).


\begin{table}[ht!]
  \centering
  \begin{tabular}{p{1.2cm}llp{2.5cm} | p{1.5cm}}
    \hline
    
      Pango Lineage & Region & Labels &
	Num. Mutations S-gene/Genome &  Num. of sequences\\
      \hline	\hline	
      B.1.1.7 & UK~\cite{galloway2021emergence} &  Alpha & 8/17 & \hskip.1in 13966\\
      B.1.351  & South Africa~\cite{galloway2021emergence}  &  Beta & 9/21& \hskip.1in 1727\\
      B.1.617.2  & India~\cite{yadav2021neutralization}  &  Delta &  8/17  & \hskip.1in 7551\\
      P.1  &  Brazil~\cite{naveca2021phylogenetic} &  Gamma &  10/21 & \hskip.1in 26629\\
      B.1.427   & California~\cite{zhang2021emergence}  & Epsilon  &  3/5 & \hskip.1in 12784\\
      \hline
  \end{tabular}
  \caption{Variants information and distribution in the dataset. The
    S/Gen. column represents number of mutations on the S gene /
    entire genome.  Total number of amino acid sequences in our
    dataset is $62,657$.}
  \label{tbl_variant_information}
\end{table}

\subsection{Data Visualization}
\begin{figure}[H]
\centering
\begin{subfigure}{.2\textwidth}
  \centering
  \includegraphics[scale = 0.14] {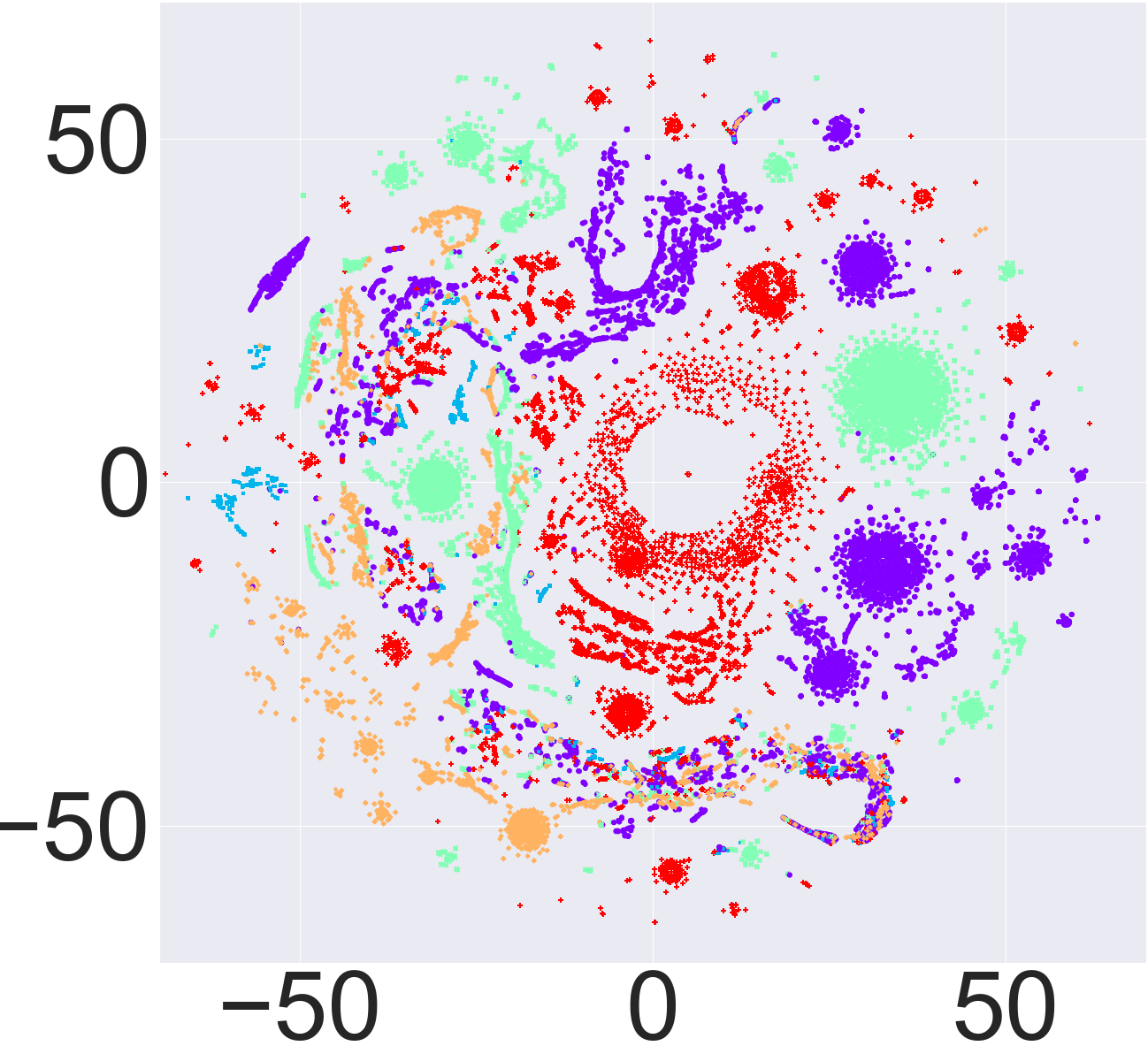}
  \caption{Original}
\end{subfigure}%
\hspace{0.7cm}
\begin{subfigure}{.2\textwidth}
  \centering
  \includegraphics[scale = 0.14] {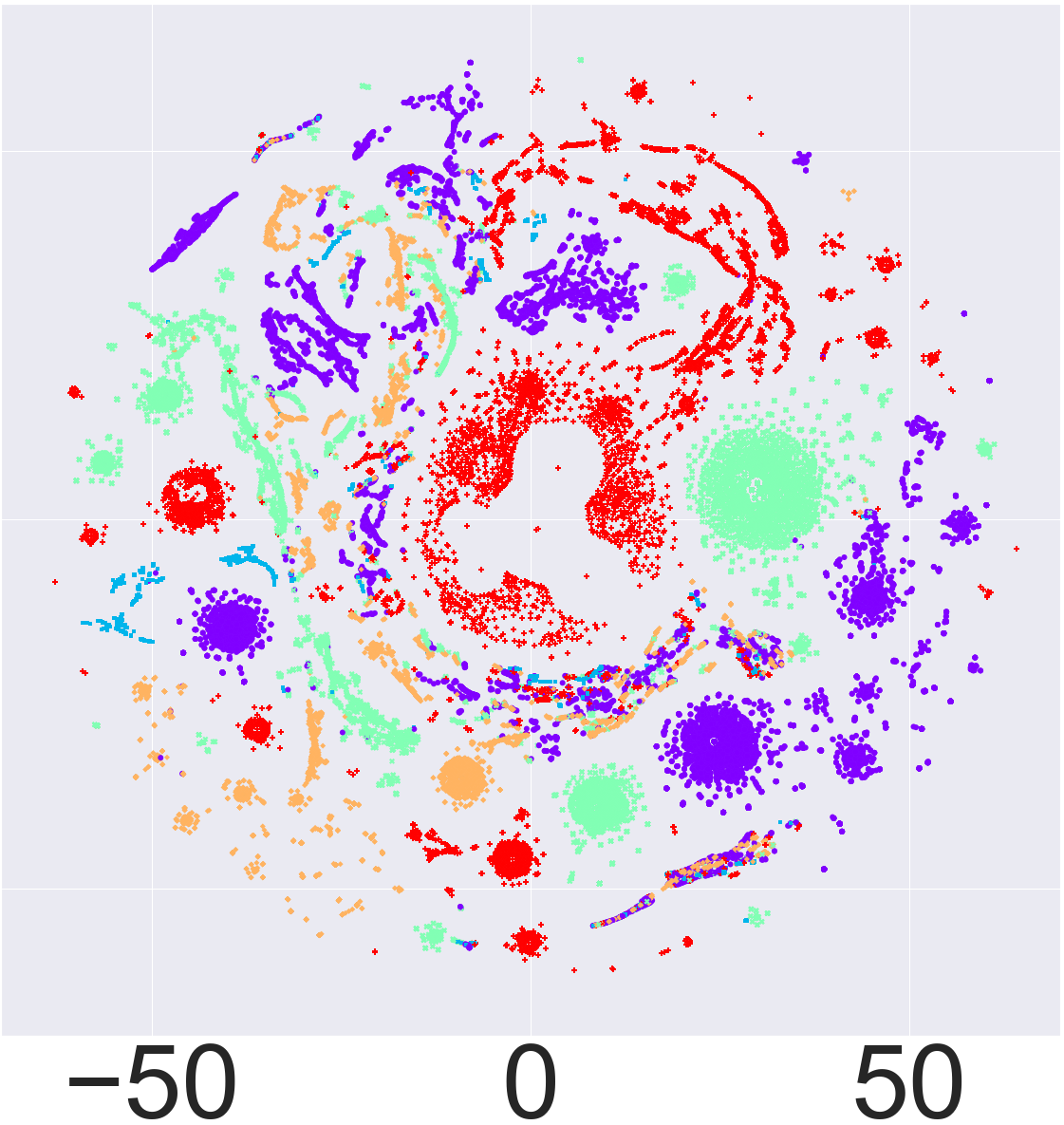}
  \caption{Boruta}
\end{subfigure}%
\hspace{0.2cm}
\begin{subfigure}{.2\textwidth}
  \centering
  \includegraphics[scale = 0.14] {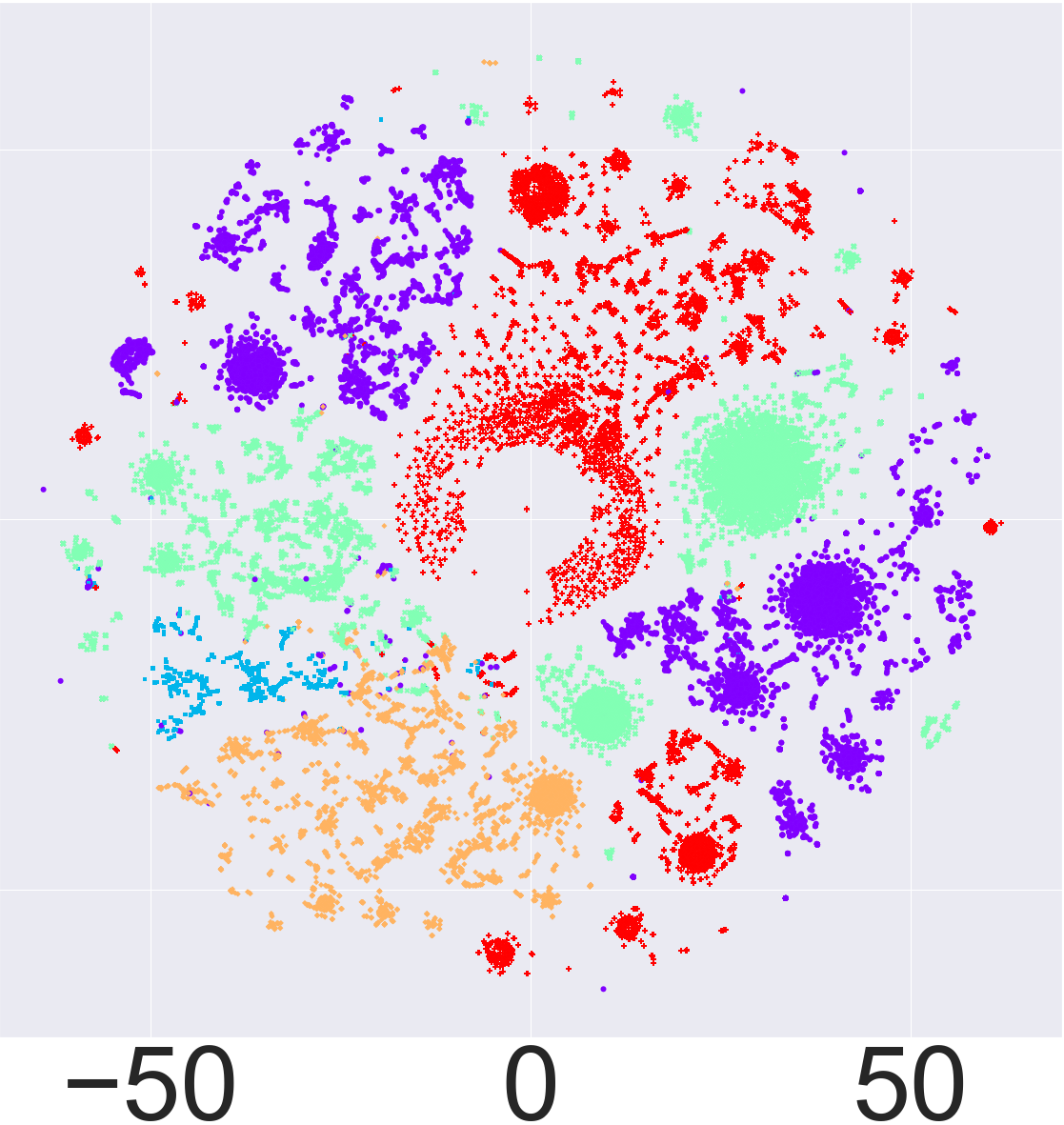}
  \caption{Lasso}
\end{subfigure}%
\hspace{0.2cm}
\begin{subfigure}{.2\textwidth}
  \centering
  \includegraphics[scale = 0.14] {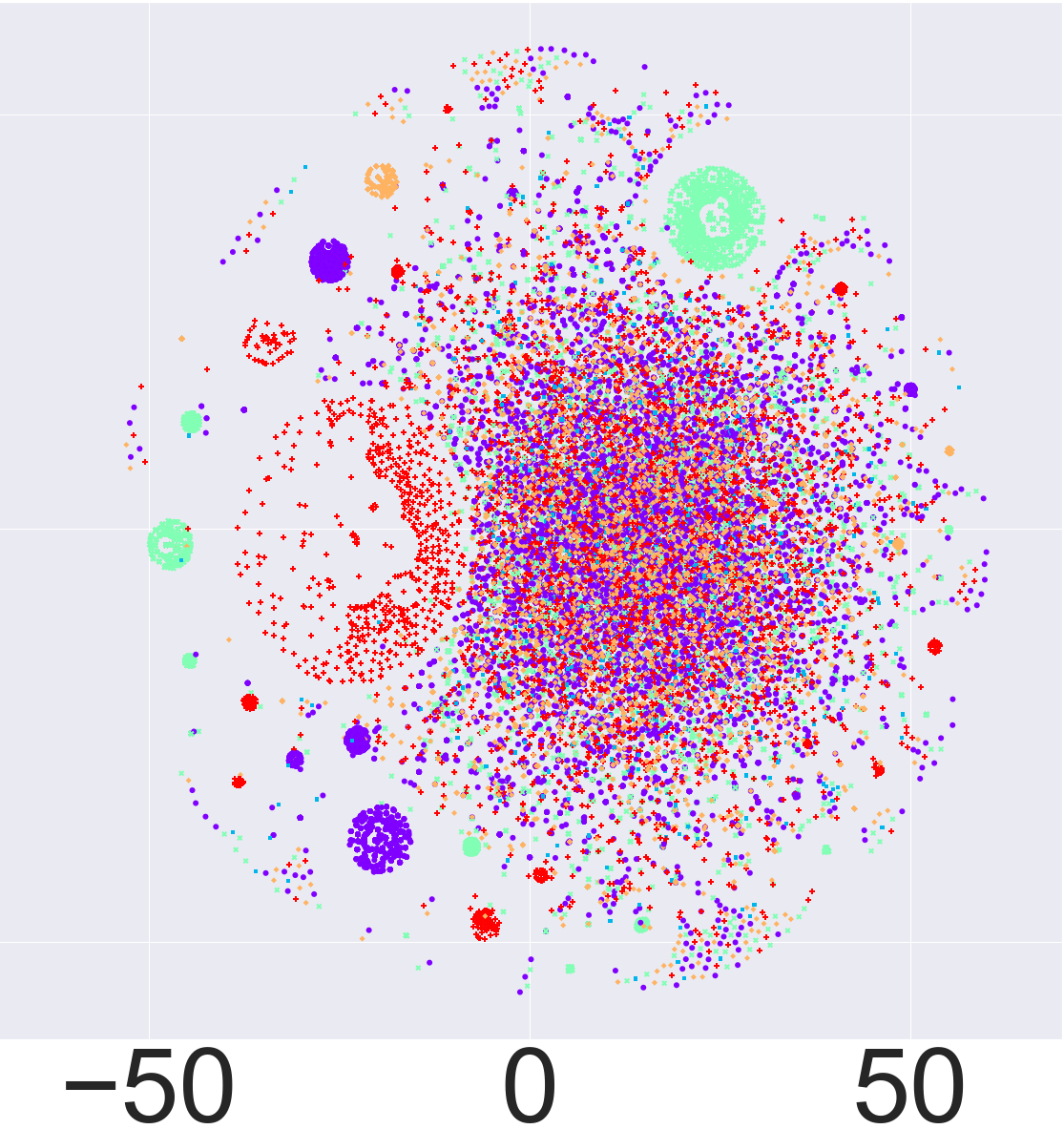}
  \caption{RFF}
\end{subfigure}%
\caption{$t$-SNE plots for original data and for different feature selection methods applied on the original data.}
\label{fig_tsne_sample}
\end{figure}

By using the $t$-SNE approach, we plotted the data to 2D real vectors to find any hidden clustering in the data. Figure~\ref{fig_tsne_sample} (a) shows the $t$-SNE plot for the GISAID dataset (before applying any feature selection). Scattered different variants everywhere is clearly visualized. Even though we cannot see clear separate clusters for each of those variants, small clusters are obvious for different variants. This evaluation for such data reveals using any clustering algorithm directly will give us good results, and some data preprocessing is curtailed for clustering the variants efficiently.

By visualizing the GISAID dataset using $t$-SNE, more clear clusters are visible after applying three different feature selection methods. In Figure~\ref{fig_tsne_sample} (b)(c)(d), we apply different feature selection methods, namely Boruta, Lasso, and RFF, respectively. We can observe that the clustering is more pure for Boruta and Lasso regression but not for RFF. This behavior shows that the supervised methods (Lasso regression and Boruta) are able to preserve the patterns in the data more effectively as compared to the unsupervised RFF.


\section{Results and Discussion}\label{sec_results_discussion}
In this section, we report the results for all clustering approaches without and with feature selection methods.
We use the weighted $F_1$ score to compute the goodness of a clustering. Since we do not have labels available for clusters, we label each cluster using the variant that have most of its sequences in that cluster (\eg, we give the label ‘Alpha’ to that cluster if most of the sequences belong to the Alpha variant).
Now, we calculate the $F_1$ score (weighted) for each cluster individually using these given labels. For different methods, the weighted $F_1$ scores are provided in Table~\ref{tbl_f1_weighted}. Note that we did not mentioned the $F_1$ scores for HDBSCAN since it is an overlapping clustering approach.
From the results, we can observe that Lasso regression is more consistent as compared to Boruta to efficiently cluster all variants. One interesting pattern we can observe is the pure clusters of some variants in case of RFF. It shows that RFF is able to cluster some variants very efficiently. However, it fails to generalize over all variants. In terms of different clustering methods, $k$-means and $k$-modes are performing better and able to generalize more on all variants as compared to the other clustering methods.


\begin{table}[ht!]
  \centering
  \begin{tabular}{lccccc}
    \hline
    & \multicolumn{5}{c}{$F_1$ Score (Weighted) for Different Variants} \\
    \cline{2-6}
    Methods & Alpha & Beta & Delta &  Gamma & Epsilon \\
    \hline	\hline	
    K-means & 0.359 & 0.157 & 0.611 & 0.690 & 0.443 \\
    K-means + Boruta & 0.418 & 0.105 & 0.610 & 0.690 & 0.652 \\
    K-means + Lasso & 0.999 & 0.007 & 0.840 & 0.999 & 0.774 \\
    K-means + RFF & 1.0 & 0.0 & 0.288 & 1.0 & 1.0 \\
    \hline
    K-modes & 0.999 & 0.005 & 0.870 & 0.998 & 0.770 \\
    K-modes + Boruta & 0.999 & 0.316 & 0.860 & 0.999 & 0.857 \\
    K-modes + Lasso & 0.999 & 0.173 & 0.917 & 0.998 & 0.076 \\
    K-modes + RFF & 1.0 & 0.0 & 0.0 & 0.613 & 1.0 \\
    \hline
    Fuzzy & 0.348 & 0.106 & 0.614 & 0.690 & 0.443 \\
    Fuzzy + Boruta & 0.357 & 0.154 & 0.613 & 0.690 & 0.443 \\
    Fuzzy + Lasso & 0.999 & 0.314 & 0.647 & 0.999 & 0.816 \\
    Fuzzy + RFF & 0.439 & 0.0 & 0.0 & 1.0 & 0.0 \\
    \hline
    Hierarchical & 0.320 & 0.103 & 0.582 & 0.704 & 0.465 \\
    Hierarchical + Boruta & 0.365 & 0.136 & 0.633 & 0.681 & 0.457 \\
    Hierarchical + Lasso & 0.995 & 0.580 & 0.578 & 0.999 & 0.834 \\
    Hierarchical + RFF & 1.0 & 0.0 & 0.288 & 1.0 & 1.0 \\
    \hline
  \end{tabular}
  \caption{Variant-wise $F_1$ (weighted) score for different clustering
    methods.}
  \label{tbl_f1_weighted}
\end{table}

\subsection{Contingency Tables}
After evaluating the clustering methods using weighted $F_1$ scores, we compute the contingency tables for variants versus clusters for different clustering approaches. The contingency tables for different clustering methods and feature selection approaches is given in Table~\ref{tbl_contingency_kmeans_kmodes} to Tables~\ref{tbl_contingency_fuzzy_hierarchy_boruta}. In Table\ref{tbl_contingency_kmeans_kmodes}, we can observe that $k$-modes without applying any feature selection is outperforming $k$-means and also the other two clustering algorithms from Table~\ref{tbl_contingency_fuzzy_hierarchy}. In Table~\ref{tbl_contingency_kmeans_kmodes_rff} and Table~\ref{tbl_contingency_fuzzy_hierarchy_rff}, we can observe that RFF is giving pure clusters for some of the variants but performing poor on the other variants. Lasso regression in Table~\ref{tbl_contingency_kmeans_kmodes_lasso} and Table~\ref{tbl_contingency_fuzzy_hierarchy_lasso}, we can observe that clusters started to become pure immediately when we apply lasso regression. This shows the effectiveness of this feature selection method for the clustering of spike sequences.
Similarly, in Table~\ref{tbl_contingency_kmeans_kmodes_boruta} and Table~\ref{tbl_contingency_fuzzy_hierarchy_boruta}, we can see that Boruta is not giving many pure clusters (apart from $k$-modes). This shows that Boruta fails to generalize over different clustering approaches and different variants.



\begin{table}[ht!]
  \centering
  \begin{tabular}{cccccc|ccccc}
    \hline
    & \multicolumn{5}{c}{K-means (Cluster IDs)} & \multicolumn{5}{c}{K-modes (Cluster IDs)} \\
    \cline{2-6} \cline{7-11}
    Variant & 0 & 1 & 2 & 3 & 4 & 0 & 1 & 2 & 3 & 4 \\
    \hline	\hline

Alpha & 1512 & 8762 & 2926 & 680 & 86  & 8 & 11492 & 284 & 330 & 1852 \\
Beta & 295 & 601 & 626 & 172 & 33 & 64 & 9 & 1604 & 31 & 19 \\
Epsilon & 956 & 7848 & 3155 & 638 & 187 & 0 & 1 & 8532 & 613 & 3638 \\
Delta & 2706 & 2605 & 1342 & 868 & 30 & 0 & 1 & 3192 & 3491 & 867 \\
Gamma & 682 & 22140 & 3016 & 741 & 50 & 26519 & 7 & 7 & 61 & 35 \\

    \hline
  \end{tabular}
  \caption{Contingency tables of variants vs clusters (No Feature Selection).}
  \label{tbl_contingency_kmeans_kmodes}
\end{table}

\begin{table}[ht!]
  \centering
  \begin{tabular}{cccccc|ccccc}
    \hline
    & \multicolumn{5}{c}{Fuzzy (Cluster IDs)} & \multicolumn{5}{c}{Hierarchical (Cluster IDs)} \\
    \cline{2-6} \cline{7-11}
    Variant & 0 & 1 & 2 & 3 & 4 & 0 & 1 & 2 & 3 & 4 \\
    \hline	\hline

Alpha & 666 & 1515 & 78 & 2945 & 8762  & 1772 & 3442 & 650 & 8036 & 66 \\
Beta & 171 & 279 & 31 & 627 & 601  & 501 & 491 & 164 & 544 & 27 \\
Epsilon & 637 & 942 & 186 & 3172 & 7847  & 1166 & 3804 & 636 & 6994 & 184 \\
Delta & 839 & 2725 & 28 & 1354 & 2605  & 2997 & 1292 & 827 & 2411 & 24 \\
Gamma & 739 & 669 & 47 & 3034 & 22140 & 865 & 3501 & 734 & 21484 & 45 \\

    \hline
  \end{tabular}
  \caption{Contingency tables of variants vs clusters (No Feature Selection).}
  \label{tbl_contingency_fuzzy_hierarchy}
\end{table}

\begin{table}[ht!]
  \centering
  \begin{tabular}{cccccc|ccccc}
    \hline
    & \multicolumn{5}{c}{K-means (Cluster IDs)} & \multicolumn{5}{c}{K-modes (Cluster IDs)} \\
    \cline{2-6} \cline{7-11}
    Variant & 0 & 1 & 2 & 3 & 4 & 0 & 1 & 2 & 3 & 4 \\
    \hline	\hline

Alpha & 0 & 12603 & 0 & 1363 & 0  & 12603 & 0 & 0 & 1363 & 0 \\
Beta & 0 & 1727 & 0 & 0 & 0  & 1727 & 0 & 0 & 0 & 0 \\
Epsilon & 0 & 10348 & 0 & 0 & 2436  & 10348 & 0 & 2436 & 0 & 0 \\
Delta & 0 & 7551 & 0 & 0 & 0  & 7551 & 0 & 0 & 0 & 0 \\
Gamma & 13076 & 12569 & 984 & 0 & 0 & 25632  & 13 & 0 & 0 & 984 \\

    \hline
  \end{tabular}
  \caption{Contingency tables of variants vs clusters (Random Fourier Transform Feature Selection).}
  \label{tbl_contingency_kmeans_kmodes_rff}
\end{table}

\begin{table}[ht!]
  \centering
  \begin{tabular}{cccccc|ccccc}
    \hline
    & \multicolumn{5}{c}{Fuzzy (Cluster IDs)} & \multicolumn{5}{c}{Hierarchical (Cluster IDs)}  \\
    \cline{2-6} \cline{7-11}
    Variant & 0 & 1 & 2 & 3 & 4  & 0 & 1 & 2 & 3 & 4 \\
    \hline	\hline

Alpha & 0 & 0 & 0 & 13966 & 0  & 12603 & 0 & 0 & 1363 & 0 \\
Beta & 0 & 0 & 0 & 1727 & 0 & 1727 & 0 & 0 & 0 & 0 \\
Epsilon & 0 & 0 & 0 & 12784 & 0  & 10348 & 0 & 2436 & 0 & 0 \\
Delta & 0 & 0 & 0 & 7551 & 0  & 7551 & 0 & 0 & 0 & 0 \\
Gamma & 0 & 0 & 0 & 13553 & 13076  & 12569 & 13076 & 0 & 0 & 984 \\

    \hline
  \end{tabular}
  \caption{Contingency tables of variants vs clusters (Random Fourier Transform Feature Selection).}
  \label{tbl_contingency_fuzzy_hierarchy_rff}
\end{table}

\begin{table}[ht!]
  \centering
  \begin{tabular}{cccccc|ccccc}
    \hline
    & \multicolumn{5}{c}{K-means (Cluster IDs)} & \multicolumn{5}{c}{K-modes (Cluster IDs)}  \\
    \cline{2-6} \cline{7-11}
    Variant & 0 & 1 & 2 & 3 & 4  & 0 & 1 & 2 & 3 & 4 \\
    \hline	\hline	
     
Alpha & 303 & 11365 & 383 & 1909 & 6  & 8 & 10958 & 282 & 2660 & 58 \\
Beta & 1551 & 4 & 148 & 23 & 1  & 65 & 9 & 1617 & 12 & 24 \\
Epsilon & 8536 & 1 & 671 & 3576 & 0  & 0 & 1 & 12000 & 112 & 671 \\
Delta & 3098 & 0 & 3693 & 760 & 0  & 0 & 0 & 3121 & 19 & 4411 \\
Gamma & 16 & 13 & 198 & 36 & 26366  & 26577 & 7& 7 & 0 & 38 \\

    \hline
  \end{tabular}
  \caption{Contingency tables of variants vs clusters (Lasso Feature Selection).}
  \label{tbl_contingency_kmeans_kmodes_lasso}
\end{table}

\begin{table}[ht!]
  \centering
  \begin{tabular}{cccccc|ccccc}
    \hline
    & \multicolumn{5}{c}{Fuzzy (Cluster IDs)} & \multicolumn{5}{c}{Hierarchical(Cluster IDs)} \\
    \cline{2-6} \cline{7-11}
    Variant & 0 & 1 & 2 & 3 & 4  & 0 & 1 & 2 & 3 & 4 \\
    \hline	\hline

Alpha & 1344 & 5 & 12042 & 362 & 213  & 1967 & 606 & 30 & 11345 & 18 \\
Beta & 99 & 1 & 6 & 440 & 1181  & 24 & 1667 & 6 & 22 & 8 \\
Epsilon & 3220 & 0 & 0 & 780 & 8784  & 3667 & 509 & 8582 & 26 & 0 \\
Delta & 4464 & 0 & 0 & 543 & 2544  & 3892 & 245 & 3367 & 40 & 7 \\
Gamma & 202 & 26169 & 16 & 232 & 10  & 12 & 1053 & 1 & 11 & 25552 \\

    \hline
  \end{tabular}
  \caption{Contingency tables of variants vs clusters (Lasso Feature Selection).}
  \label{tbl_contingency_fuzzy_hierarchy_lasso}
\end{table}

\begin{table}[ht!]
  \centering
  \begin{tabular}{cccccc|ccccc}
    \hline
    & \multicolumn{5}{c}{K-means(Cluster IDs)} & \multicolumn{5}{c}{K-modes(Cluster IDs)} \\
    \cline{2-6} \cline{7-11}
    Variant & 0 & 1 & 2 & 3 & 4  & 0 & 1 & 2 & 3 & 4 \\
    \hline	\hline

Alpha & 8762 & 86 & 2925 & 680 & 1513  & 11403 & 7 & 184 & 1823 & 549 \\
Beta & 601 & 33 & 626 & 172 & 295  & 6 & 6 & 640 & 1060 & 15 \\
Epsilon & 7848 & 187 & 3155 & 638 & 956  & 1 & 0 & 11170 & 947 & 666 \\
Delta & 2605 & 30 & 1342 & 868 & 2706  & 0 & 0 & 2894 & 690 & 3967 \\
Gamma & 22140 & 50 & 3016 & 741 & 682  & 6 & 25428 & 6 & 1128 & 61 \\

    \hline
  \end{tabular}
  \caption{Contingency tables of variants vs clusters (Boruta Feature Selection).}
  \label{tbl_contingency_kmeans_kmodes_boruta}
\end{table}

\begin{table}[ht!]
  \centering
  \begin{tabular}{cccccc|ccccc}
    \hline
    & \multicolumn{5}{c}{Fuzzy (Cluster IDs)} & \multicolumn{5}{c}{Hierarchical (Cluster IDs)} \\
    \cline{2-6} \cline{7-11}
    Variant & 0 & 1 & 2 & 3 & 4  & 0 & 1 & 2 & 3 & 4 \\
    \hline	\hline

Alpha & 668 & 1513 & 78 & 2945 & 8762  & 9373 & 702 & 2641 & 1198 & 52 \\
Beta & 171 & 297 & 31 & 627 & 601  & 823 & 170 & 457 & 254 & 23 \\
Epsilon & 637 & 943 & 186 & 3170 & 7848  & 8419 & 644 & 2949 & 591 & 181 \\
Delta & 851 & 2713 & 28 & 1354 & 2605  & 2847 & 879 & 1563 & 2245 & 17 \\
Gamma & 739 & 669 & 47 & 3034 & 22140 & 22955 & 743 & 2330 & 560 & 41 \\

    \hline
  \end{tabular}
  \caption{Contingency tables of variants vs clusters (Boruta Feature Selection).}
  \label{tbl_contingency_fuzzy_hierarchy_boruta}
\end{table}

\subsection{HDBSCAN Clustering}
\begin{figure}[ht!]
\centering
\begin{subfigure}{.5\textwidth}
  \centering
  \includegraphics[scale = 0.2] {Figures/New_Plot_orginal_data.png}
  \caption{Original Data.}
\end{subfigure}%
\begin{subfigure}{.5\textwidth}
  \centering
  \includegraphics[scale = 0.2] {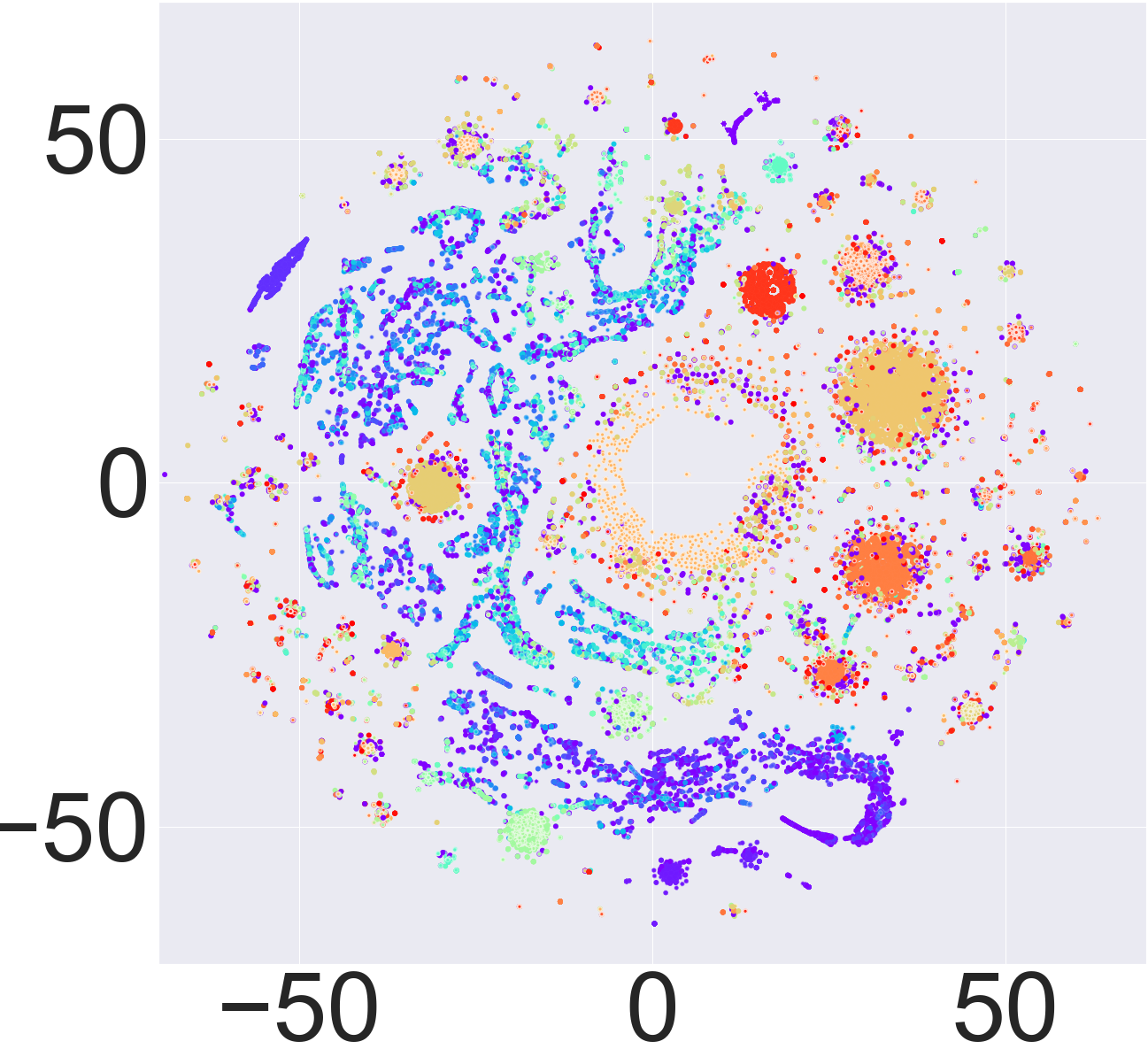}
  \caption{HDBSCAN (no feature selection).}
\end{subfigure}
\caption{(a) t-SNE plots for the original variants as labels, (b) t-SNE plot with labels got after applying HDBSCAN without any feature selection method on the frequency vectors. }
\label{fig_tsne_hdbscan_orig}
\end{figure}

\begin{figure}[ht!]
\centering
\begin{subfigure}{.2\textwidth}
  \centering
  \includegraphics[scale = 0.14] {Figures/New_HDBSCAN_org_5clusters.png}
  \caption{No Feat. Selection}
\end{subfigure}%
\hspace{0.7cm}
\begin{subfigure}{.2\textwidth}
  \centering
  \includegraphics[scale = 0.14] {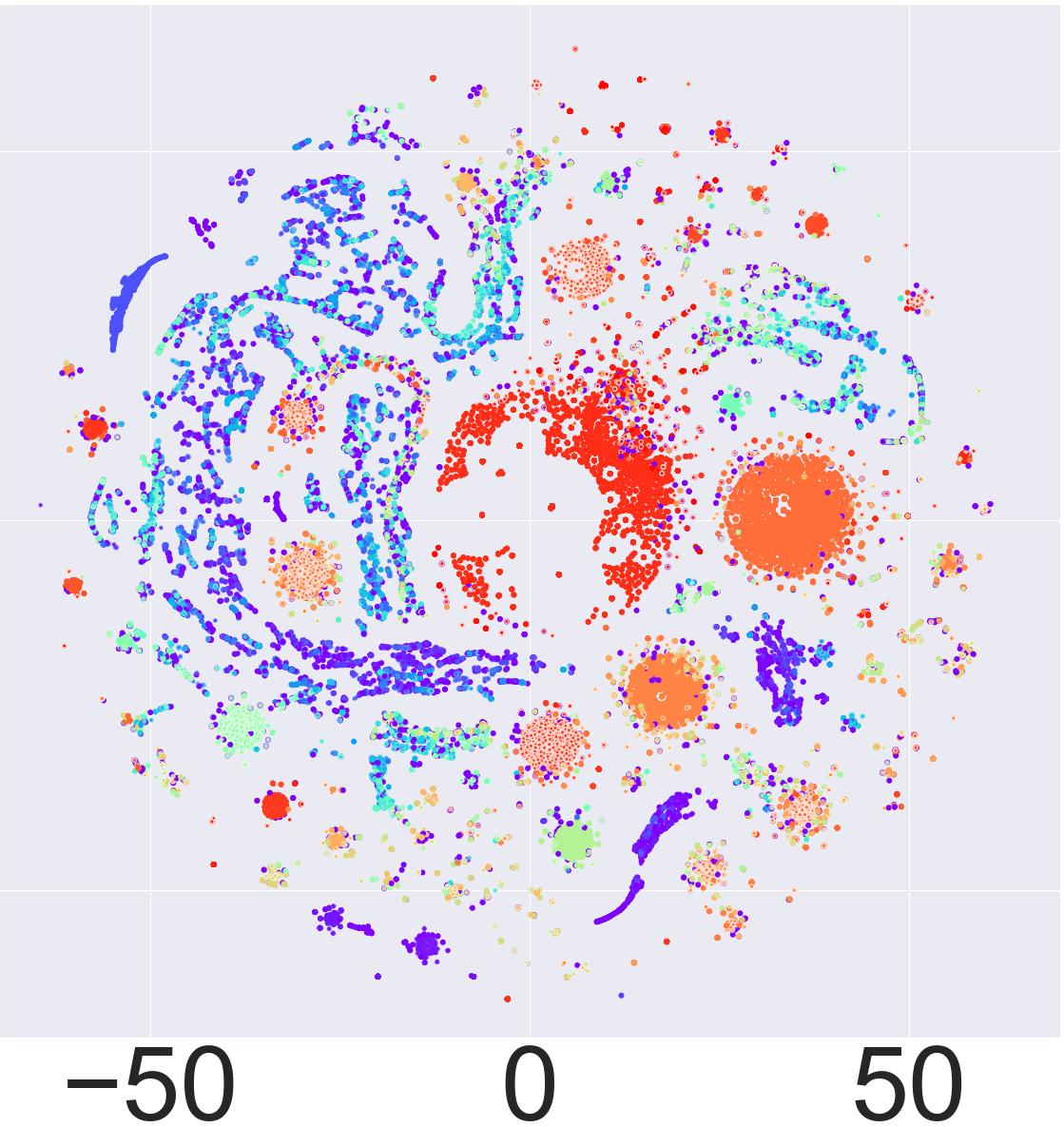}
  \caption{Boruta}
\end{subfigure}%
\hspace{0.2cm}
\begin{subfigure}{.2\textwidth}
  \centering
  \includegraphics[scale = 0.14] {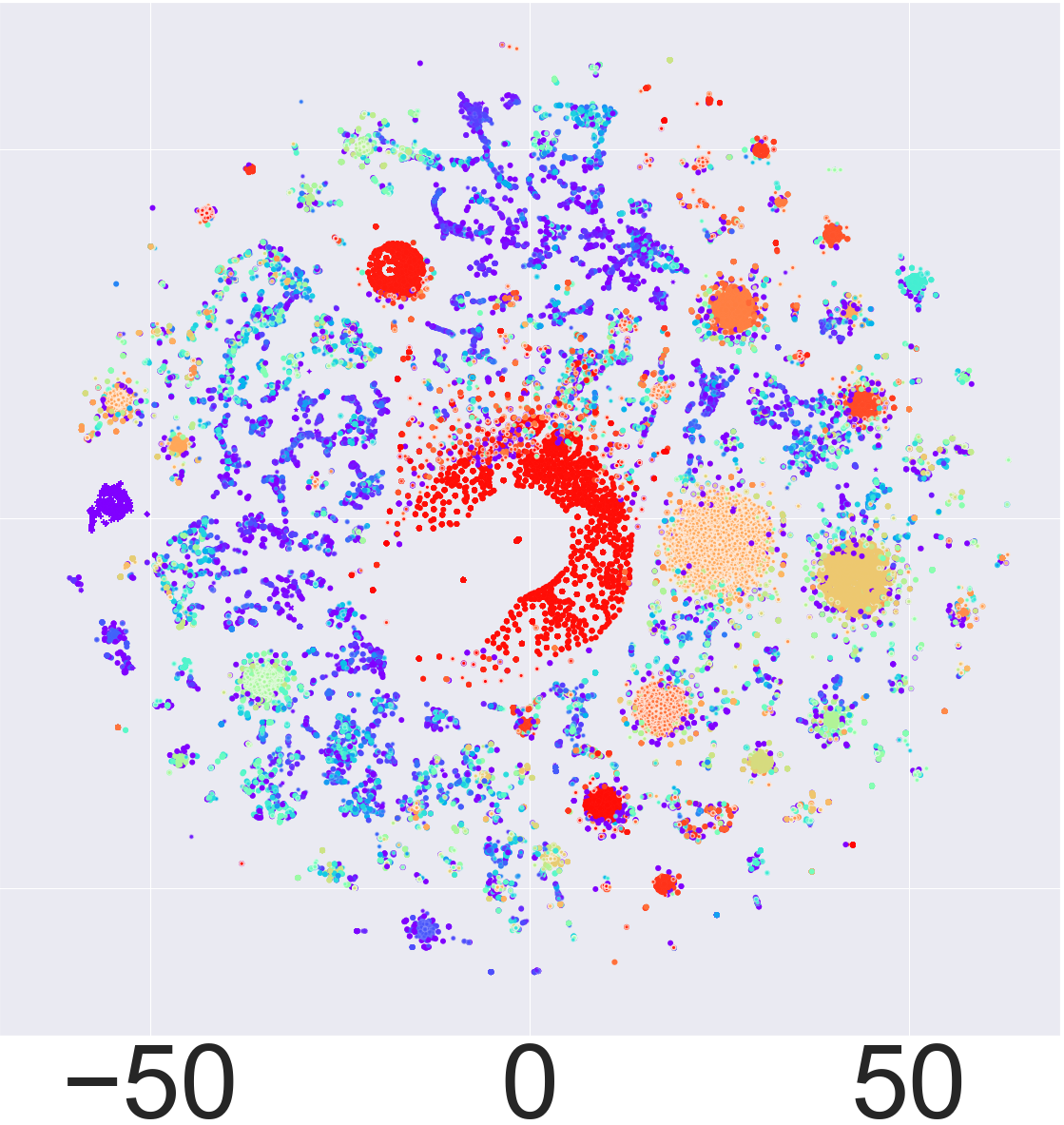}
  \caption{Lasso}
\end{subfigure}%
\hspace{0.2cm}
\begin{subfigure}{.2\textwidth}
  \centering
  \includegraphics[scale = 0.14] {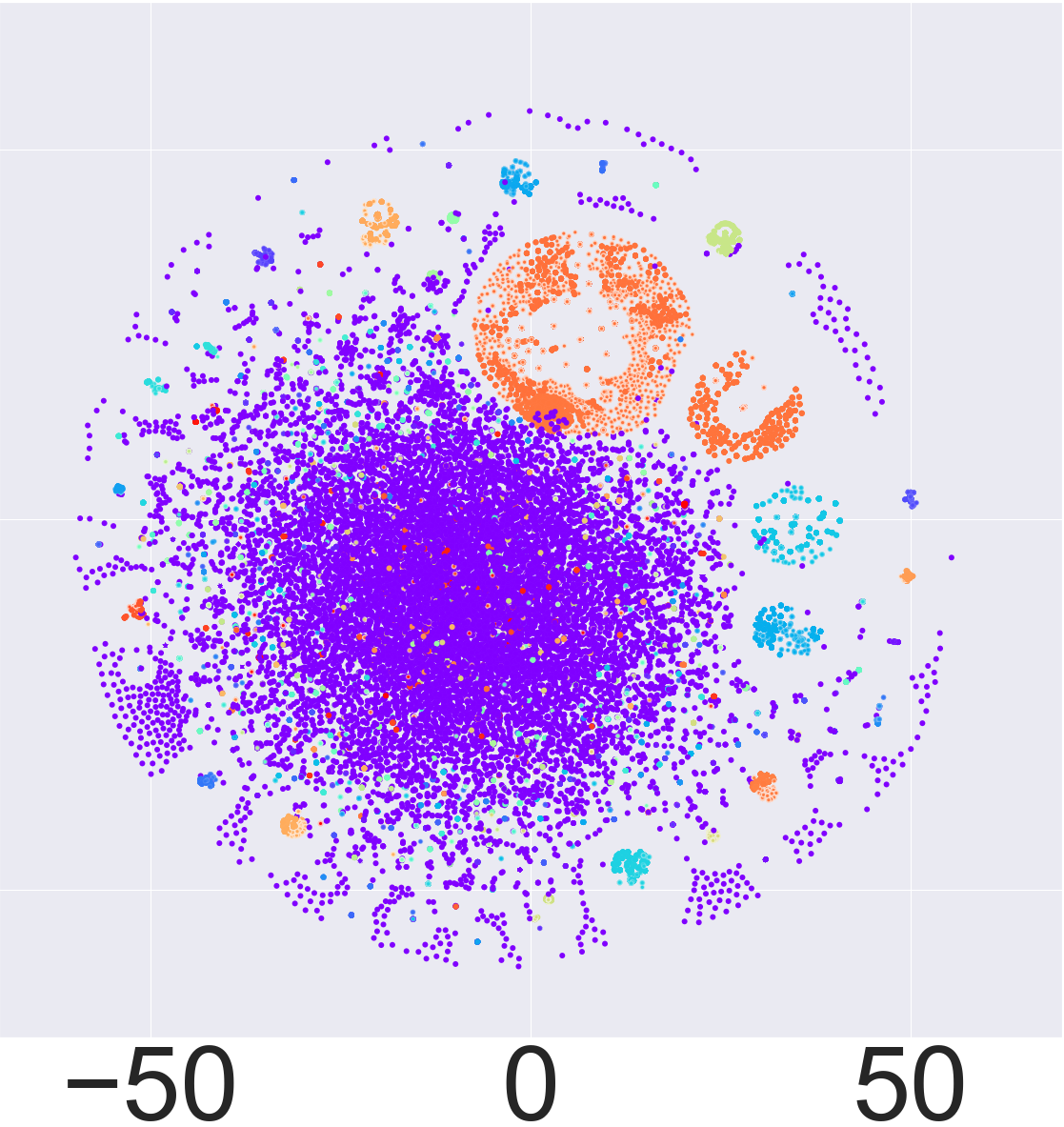}
  \caption{RFF}
\end{subfigure}%
\caption{t-SNE plots for HDBSCAN without and with feature selection methods.}
\label{fig_hdbcan_original}
\end{figure}

After doing analysis on hard clustering algorithms, we evaluate the performance of the soft clustering approach (HDBSCAN) in this section. 
To evaluate HDBSCAN, we use the $t$-SNE approach to plot the original variants from our data and compared them with clusters we obtained after applying HDBSCAN. Since this is a soft clustering approach (overlapping allowed), there were large number of clusters inferred for different feature selection methods. Therefore we use $t$-SNE to plot the clusters to visually observe the patters before and after clustering. Figure~\ref{fig_tsne_hdbscan_orig} shows the comparison on t-SNE plot on original data versus $t$-SNE plots for the clustering results after applying HDBSCAN. Since overlapping is allowed in this setting, we cannot see any pure clusters as compared to the original $t$-SNE plot. An interesting finding from such result is that not all sequences corresponding to a specific variant are similar to each other. This means that a small cluster of sequences, that initially belong to a certain variant can make another subgroup, which could eventually lead to developing a new variant. Therefore, using such overlapping clustering approach, we can visually observe if a group of sequences are diverging from its parent variant. Biologists and other decision making authorities can then take relevant measure to deal with such scenarios. The $t$-SNE plots for different feature selection methods in given in Figure~\ref{fig_hdbcan_original}.


\subsection{Runtime Comparison}
\begin{figure}[ht!]
    \centering
    \includegraphics[scale = 0.3]{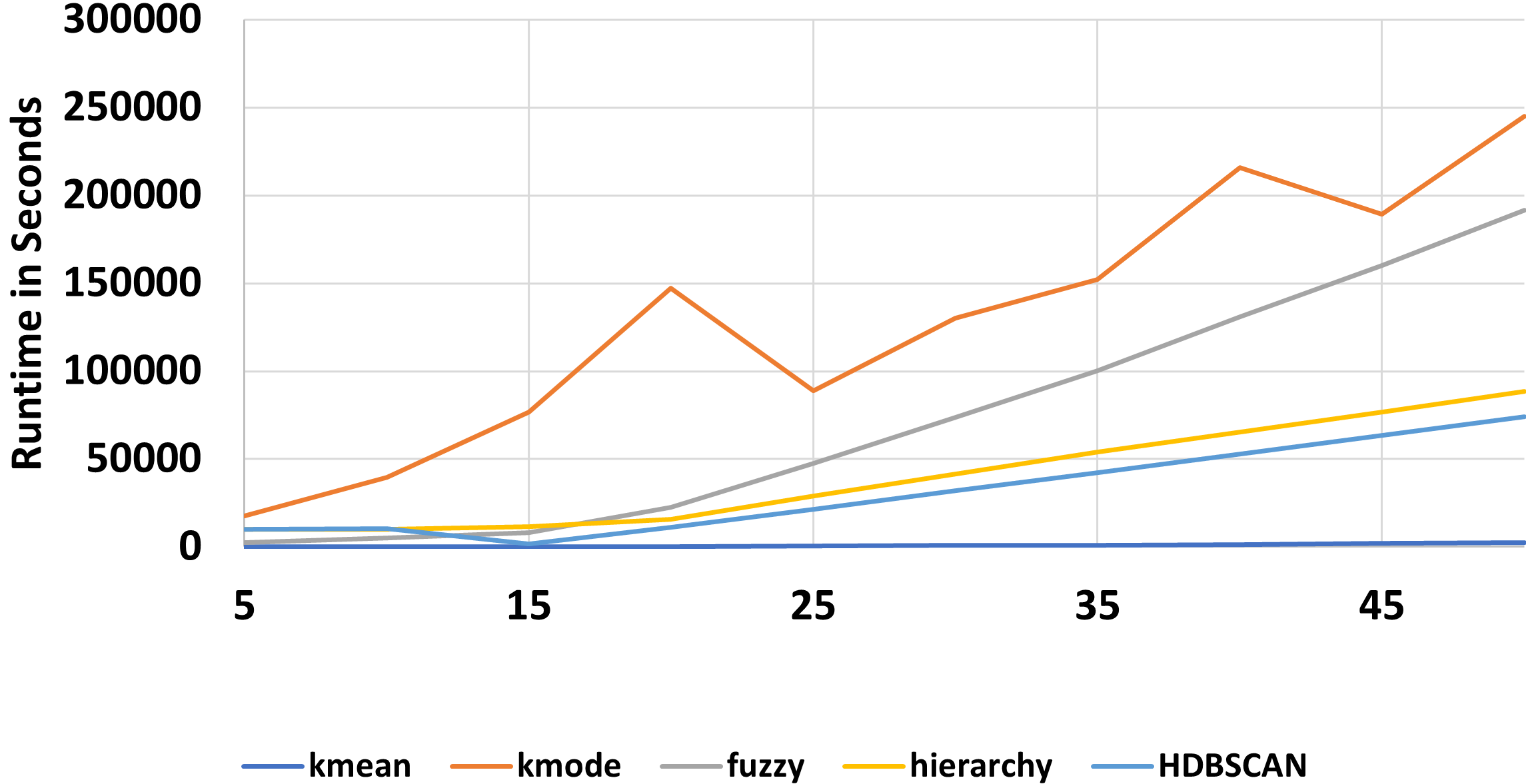}
    \caption{Running time for different clustering methods (No feature selection method). X-axis shows number of clusters.}
    \label{fig_runtime_no_feat_selection}
\end{figure}

\begin{figure}[ht!]
    \centering
    \includegraphics[scale = 0.3]{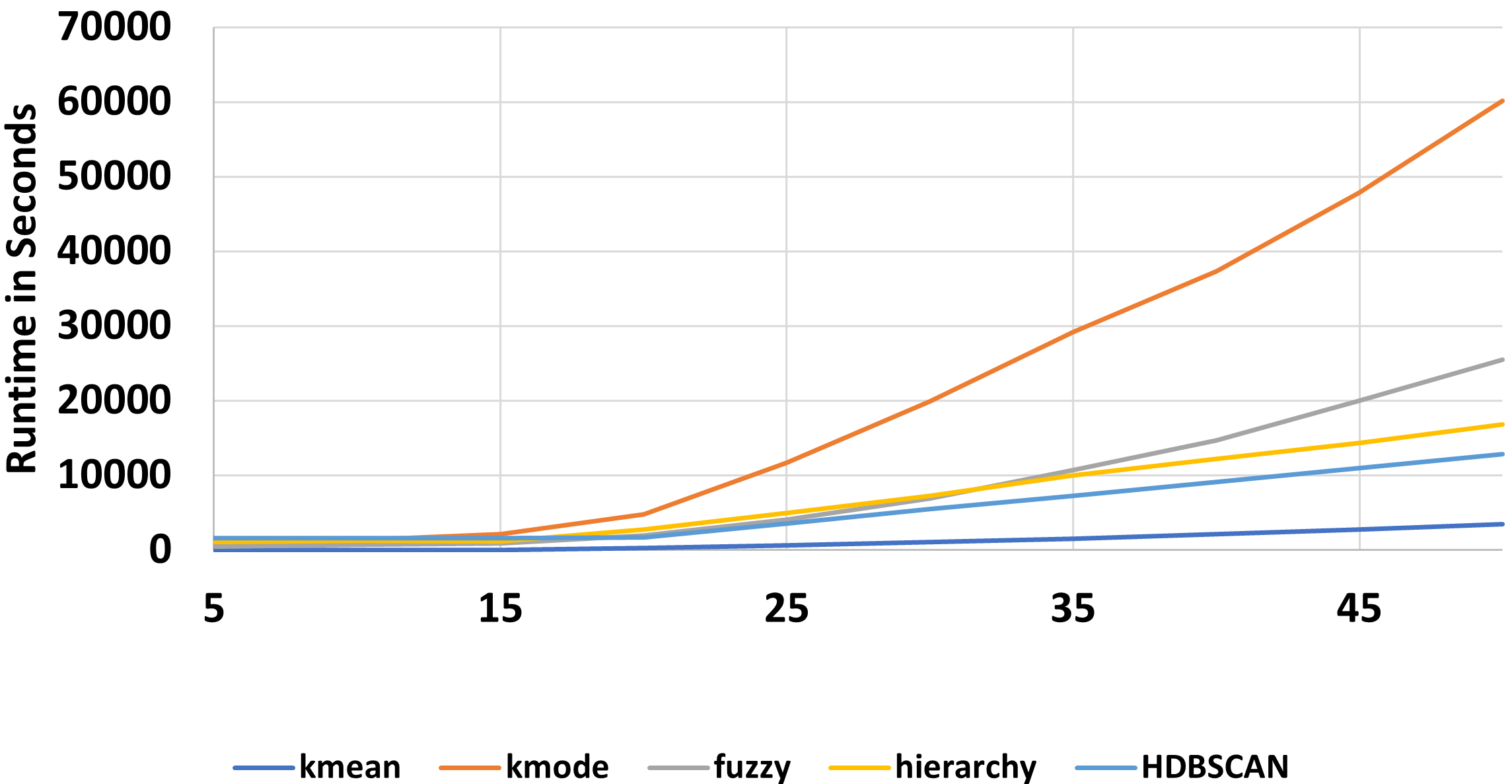}
    \caption{Running time for different clustering methods (Random Fourier Transform Feature Selection). X-axis shows number of clusters.}
    \label{fig_runtime_rff}
\end{figure}

\begin{figure}[ht!]
    \centering
    \includegraphics[scale = 0.3]{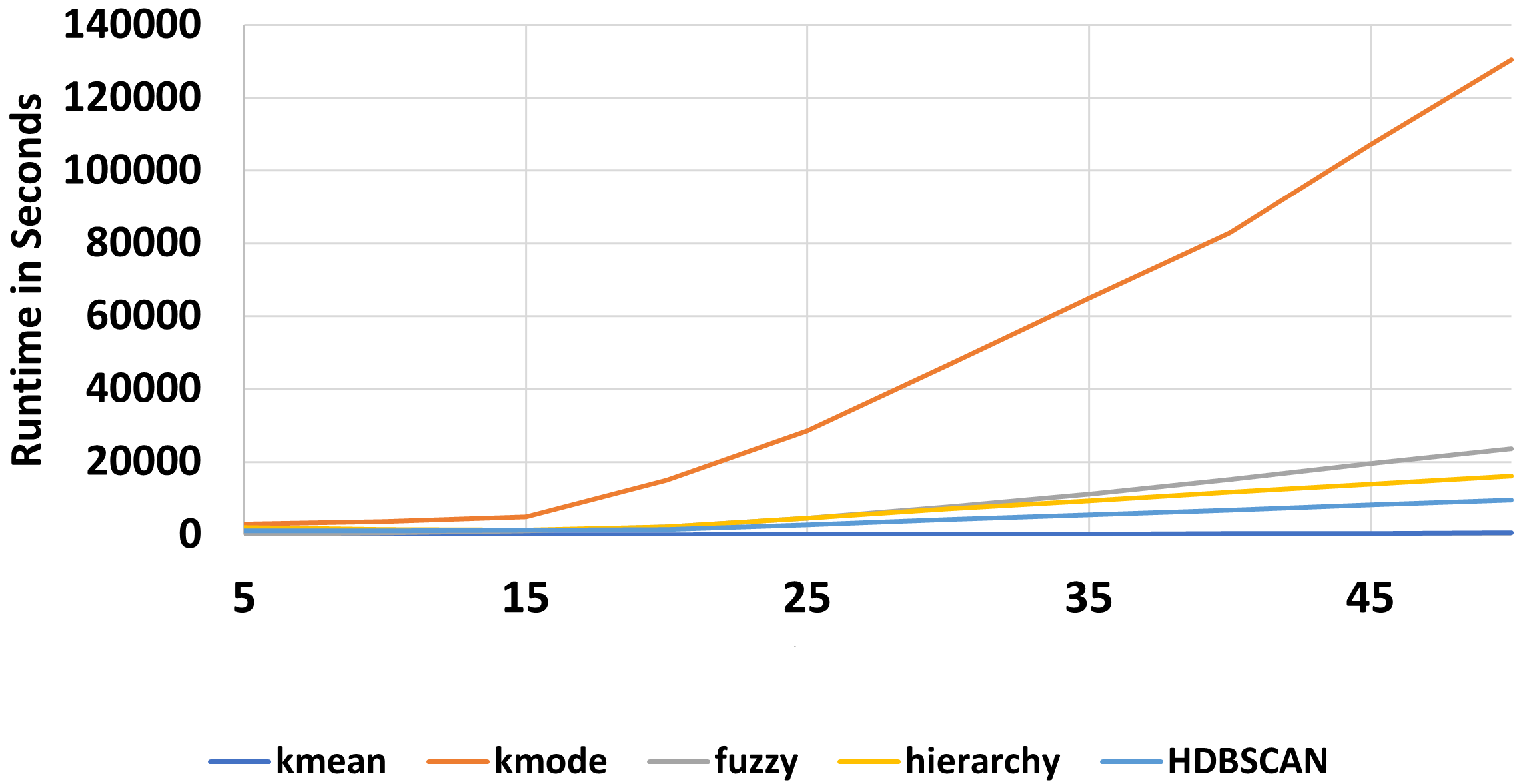}
    \caption{Running time for different clustering methods (Boruta Feature Selection). X-axis shows number of clusters.}
    \label{fig_runtime_boruta}
\end{figure}

\begin{figure}[ht!]
    \centering
    \includegraphics[scale = 0.3]{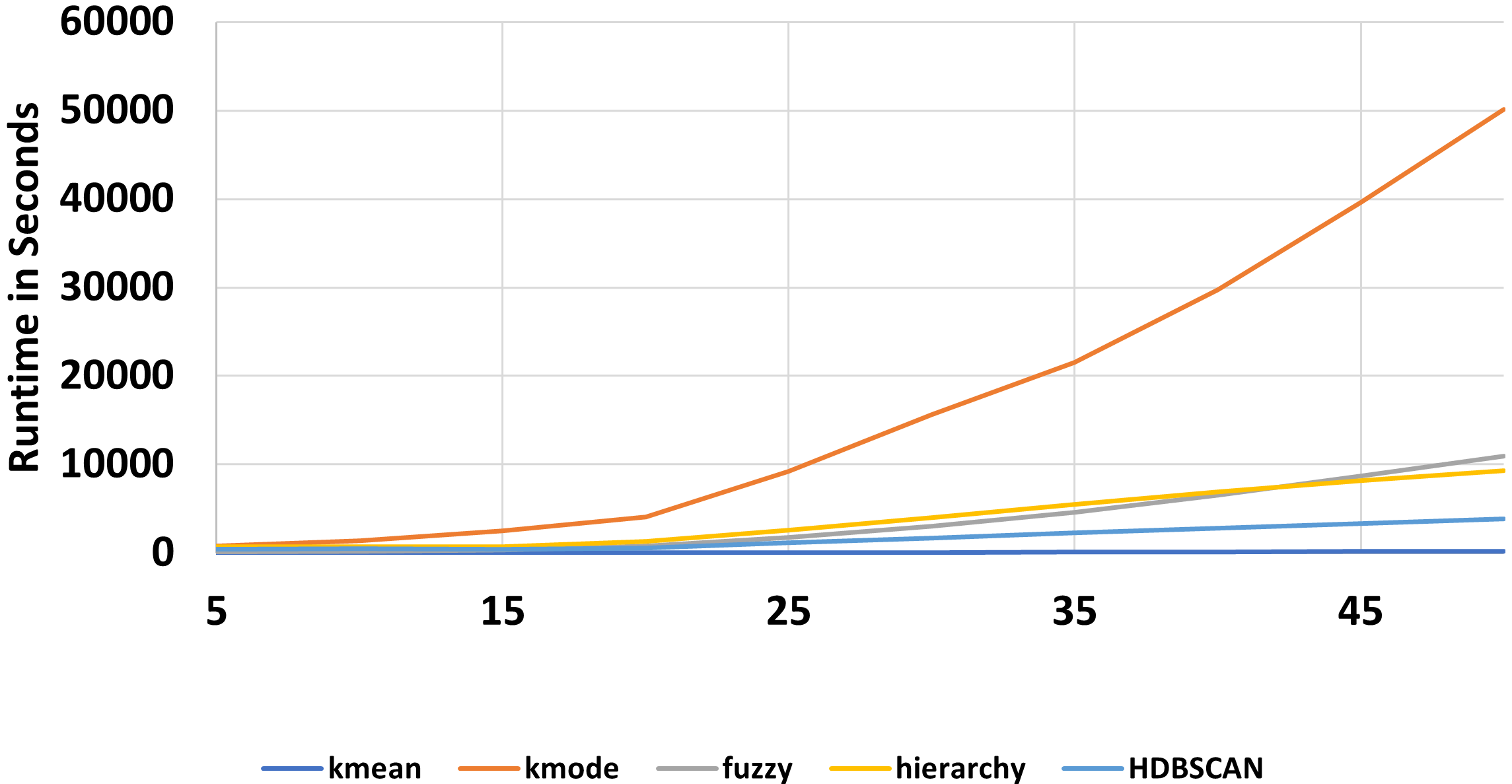}
    \caption{Running time for different clustering methods (Lasso Feature Selection). X-axis shows number of clusters.}
    \label{fig_runtime_lasso}
\end{figure}

After applying different clustering methods and feature selection algorithms on the spike sequences, we observe that $k$-means and $k$-modes are performing better than the other clustering methods in terms of weighted $F_1$ score. However, it is also important to study the effect of runtime for these clustering approaches so that we can evaluate the trade-off between $F_1$ score and the runtime. For this purpose, we compute the runtime of different clustering algorithms without and with feature selection methods. Figure~\ref{fig_runtime_no_feat_selection} shows the runtime for all five clustering methods without applying any feature selection on the data. We can observe that $k$-modes is very expensive in terms of runtime and $k$-means takes the least amount of time to execute. Similar behavior is observed in Figure~\ref{fig_runtime_rff}, Figure~\ref{fig_runtime_boruta}, and Figure~\ref{fig_runtime_lasso} for RFF, Boruta, and Lasso regression, respectively. This behavior shows that although $k$-modes is performing better in terms of $F_1$ score, it is an outlier in terms of runtime. This behavior also shows the effectiveness of the $k$-means algorithm in terms of $F_1$ score and also in terms of runtime.

\section{Conclusion}
\label{sec_conclusion}

We propose a feature vector representation and a set of feature
selection methods to eliminate the less important features, allowing
many different clustering methods to successfully cluster SARS-CoV-2
spike protein sequences with high $F_1$ scores.  We show that runtime
is also an important factor while clustering the coronavirus spike
sequences. The $k$-means algorithm is able to generalize over all
variants in terms of doing pure clustering and also consuming the
least amount of runtime.  One possible future work is to use more data
for the analysis.  Testing out additional clustering methods could be
another direction.  Using deep learning on even bigger data could give
us some interesting insights. Another interesting extension is to
compute other feature vector representations, \eg, based on
minimizers, which can be done without the need for aligning the
sequences.  This would allow us to use all of this clustering
machinery to study unaligned (even unassembled) sequencing reads of
intra-host viral populations, to unveil the interesting dynamics at
this scale.

\reftitle{References}
\externalbibliography{yes}
\bibliography{clustering_covid}


\end{document}